\documentclass[conference]{IEEEtran}
\IEEEoverridecommandlockouts

\usepackage{amsmath,amssymb,amsfonts}
\usepackage{mathtools}
\DeclareMathOperator{\sgn}{sgn}

\usepackage[acronym,indexonlyfirst,nomain]{glossaries}
\newacronym[plural={SVMs}, longplural={Support Vector Machines}]{svm}{SVM}{Support Vector Machine}
\newacronym[plural={LS-SVMs}, longplural={Least Squares Support Vector Machines}]{lssvm}{LS-SVM}{Least Squares Support Vector Machine}
\newacronym{plssvm}{PLSSVM}{Parallel Least Squares Support Vector Machine}

\newacronym[plural={GPGPUs}, longplural={General Purpose Graphics Processing Units}]{gpgpu}{GPGPU}{General Purpose Graphics Processing Unit}
\newacronym[plural={GPUs}, longplural={Graphics Processing Units}]{gpu}{GPU}{Graphics Processing Unit}
\newacronym[plural={CPUs}, longplural={Central Processing Units}]{cpu}{CPU}{Central Processing Unit}

\newacronym{smo}{SMO}{Sequential Minimal Optimization}

\newacronym{hpc}{HPC}{High Performance Computing}

\newacronym{cuda}{CUDA}{Compute Unified Device Architecture}
\newacronym{opencl}{OpenCL}{Open Computing Language}
\newacronym{mpi}{MPI}{Message Passing Interface}

\newacronym{soa}{SoA}{Structure-of-Arrays}
\newacronym{aos}{AoS}{Array-of-Structures}

\newacronym{cg}{CG}{Conjugate Gradients}

\newacronym{raii}{RAII}{Resource Acquisition Is Initialization}

\newacronym[plural={SMs}, longplural={Streaming Multiprocessors}]{sm}{SM}{Streaming Multiprocessor}

\usepackage[binary-units=true]{siunitx}
\DeclareSIUnit\flops{FLOPS}
\def\BibTeX{{\rm B\kern-.05em{\sc i\kern-.025em b}\kern-.08em
		T\kern-.1667em\lower.7ex\hbox{E}\kern-.125emX}}
\usepackage[numbers, sort]{natbib}
\usepackage{threeparttable}
\usepackage{makecell}

\usepackage{algorithmicx}
\usepackage{algorithm}
\usepackage{algpseudocode}

\ifCLASSOPTIONcompsoc
    \usepackage[caption=false, font=normalsize, labelfont=sf, textfont=sf]{subfig}
\else
	\usepackage[caption=false, font=footnotesize]{subfig}
\fi

\usepackage{listings}
\usepackage{csquotes}
\usepackage{graphicx}
\usepackage{pifont}
\usepackage{textcomp}
\usepackage[dvipsnames]{xcolor}

\usepackage[disable]{todonotes}

\renewcommand{\equationautorefname}{Equation}
\usepackage{subfig}
\renewcommand{\figureautorefname}{Figure}
\newcommand{\subfigureautorefname}{Figure}
\renewcommand{\sectionautorefname}{Section}

\newcommand{\Autoref}[1]{%
  \begingroup%
  \renewcommand\equationautorefname{Equation}%
  \renewcommand{\figureautorefname}{Figure}
  \renewcommand{\subfigureautorefname}{Figure}%
  \renewcommand{\sectionautorefname}{Section}%
  \autoref{#1}%
  \endgroup%
}

\begin{document}

\title{PLSSVM: A (multi-)GPGPU-accelerated Least Squares Support Vector Machine}

\author{\IEEEauthorblockN{1\textsuperscript{st} Alexander Van Craen}
	\IEEEauthorblockA{\textit{IPVS} \\
		\textit{University of Stuttgart}\\
		Stuttgart, Germany \\
		Alexander.Van-Craen@ipvs.uni-stuttgart.de}
	\and
	\IEEEauthorblockN{2\textsuperscript{nd} Marcel Breyer}
	\IEEEauthorblockA{\textit{IPVS} \\
		\textit{University of Stuttgart}\\
		Stuttgart, Germany \\
		Marcel.Breyer@ipvs.uni-stuttgart.de}
	\and
	\IEEEauthorblockN{3\textsuperscript{rd} Dirk Pflüger}
	\IEEEauthorblockA{\textit{IPVS} \\
		\textit{University of Stuttgart}\\
		Stuttgart, Germany \\
		Dirk.Pflueger@ipvs.uni-stuttgart.de}
}
\makeatletter
\def\ps@IEEEtitlepagestyle{%
  \def\@oddfoot{\mycopyrightnotice}%
  \def\@oddhead{\hbox{}\@IEEEheaderstyle\leftmark\hfil\thepage}\relax
  \def\@evenhead{\@IEEEheaderstyle\thepage\hfil\leftmark\hbox{}}\relax
  \def\@evenfoot{}%
}
\def\mycopyrightnotice{%
  \begin{minipage}{\textwidth}
  \centering \scriptsize
  \copyright~2022 IEEE.  Personal use of this material is permitted.  Permission from IEEE must be obtained for all other uses, in any current or future media, including reprinting/republishing this material for advertising or promotional purposes, creating new collective works, for resale or redistribution to servers or lists, or reuse of any copyrighted component of this work in other works (doi: \href{http://dx.doi.org/10.1109/IPDPSW55747.2022.00138}{10.1109/IPDPSW55747.2022.00138}). \hfill
  \end{minipage}
}
\makeatother
\maketitle

\begin{abstract}
    Machine learning algorithms must be able to efficiently cope with massive data sets.
    Therefore, they have to scale well on any modern system and be able to exploit the computing power of accelerators independent of their vendor.
    In the field of supervised learning, \glspl{svm} are widely used.
    However, even modern and optimized implementations such as LIBSVM or ThunderSVM do not scale well for large non-trivial dense data sets on cutting-edge hardware:
    Most \acrshort{svm} implementations are based on \acrlong{smo}, an optimized though inherent sequential algorithm. 
    Hence, they are not well-suited for highly parallel \acrshortpl{gpu}. 
    Furthermore, we are not aware of a performance portable implementation that supports \acrshortpl{cpu} and \acrshortpl{gpu} from different vendors.

    We have developed the \acrshort{plssvm} library to solve both issues. 
    First, we resort to the formulation of the \acrshort{svm} as a least squares problem. 
    Training an \acrshort{svm} then boils down to solving a system of linear equations for which highly parallel algorithms are known. 
    Second, we provide a hardware independent yet efficient implementation: \acrshort{plssvm} uses different interchangeable backends---OpenMP, \acrshort{cuda}, \acrshort{opencl}, SYCL---supporting modern hardware from various vendors like NVIDIA, AMD, or Intel on multiple \acrshortpl{gpu}. 
    \acrshort{plssvm} can be used as a drop-in replacement for LIBSVM. 
    We observe a speedup on \acrshortpl{cpu} of up to $10$ compared to LIBSVM and on \acrshortpl{gpu} of up to $14$ compared to ThunderSVM. 
    Our implementation scales on many-core \acrshortpl{cpu} with a parallel speedup of $\num{74.7}$ on up to $256$ \acrshort{cpu} threads and on multiple \acrshortpl{gpu} with a parallel speedup of $\num{3.71}$ on four \acrshortpl{gpu}.
    
    The code, utility scripts, and documentation are all available on GitHub:  \url{https://github.com/SC-SGS/PLSSVM}.
\end{abstract}

\glsreset{svm}



\section{Introduction}
\label{sec:intro}

The two most common tasks in supervised machine learning are classification and regression.
Classification predicts to which set of categories/classes objects or situations belong,
while regression estimates the function value of a functional dependency, e.g., a statistical process.
Both can be solved using the same algorithms, by matching a function value to two or more discrete classes.
Moreover, both share the challenge to scale in the size of the data in order to cope with massive data sets.

In the classification task, the well-known and commonly used \glspl{svm} show competitive performance and have very few parameters to tune.
The widely used \gls{svm} library LIBSVM~\cite{CC01a} has already been cited over $\num{50000}$ times.
A survey conducted by Kaggle in 2017 showed that $\SI{26}{\percent}$ of the data mining and machine learning practitioners use \glspl{svm} \cite{kaggle2017survey}.
Examples of current research topics are COVID-19 pandemic predictions~\cite{singh2020prediction}, automatic COVID-19 lung image classification~\cite{mahdy2020automatic}, forecasting of carbon prices~\cite{Bangzhu2022Forecasting}, face detection~\cite{riyazuddin2020effective}, or propaganda text recognition~\cite{Khanday2021propaganda}.

The basic idea of the \glspl{svm} was introduced in 1992 by \citet{Boser:1992:TAO:130385.130401} and extended in 1995 by \citet{Cortes1995} to the today's state-of-the-art approach.
Based on this approach, \citeauthor{Joachims/02a} published one of the first \gls{svm} libraries SVM\textsuperscript{light}~\cite{Joachims/02a}.
\citeauthor{CC01a} developed LIBSVM~\cite{CC01a}, which is one of the most widely used \gls{svm} libraries today.
Both are based on the \gls{smo} algorithm \cite{platt1998sequential}.
Additionally, plenty of research have focused on parallelizing \glspl{svm} on multi-core \gls{cpu} systems. 
For example, \citet{fan2008liblinear} developed LIBLINEAR as a LIBSVM extension and \citet{zeng2008fast} described a novel parallel \gls{smo} algorithm.

ThunderSVM~\cite{wenthundersvm18} is a rather new \gls{svm} implementation supporting \glspl{cpu} and NVIDIA \glspl{gpu} using the \gls{cuda}.
Additionally, there have been various other attempts to port \glspl{svm} efficiently to \glspl{gpu}, e.g., with CUDA~\cite{carpenter2009cusvm,li2011gpusvm,sergherrero2015multisvm}, but also using other languages and frameworks such as the \gls{opencl}~\cite{Cagnin2015}, Python~\cite{raschka2020machine,murtazajafferji2018svmgpu}, OpenACC~\cite{Codreanu2014}, or SYCL~\cite{2020sycl-ml}.
However, all approaches mentioned so far are based on \gls{smo}, an inherently sequential algorithm, and, therefore, not very well suited for high performance \gls{gpu} implementations and vast data sets.

\citeauthor{Suykens1999}~\cite{Suykens1999, suykens2002least} developed the least squares formulation of the \gls{svm} problem.
In this formulation, training an \gls{svm} is reduced to solving a system of linear equations, a problem for which highly parallel algorithms exist.
\citeauthor{suykens2002weighted} improved their \gls{lssvm} further by introducing weights \cite{suykens2002weighted} and extending their implementation to also support sparse data structures \cite{suykens2000sparse}.
Furthermore, they introduced the multi-class classification for \glspl{lssvm} \cite{suykens1999multiclass}.
An exact correlation between \glspl{svm} using \gls{smo} and \glspl{lssvm} was investigated by \citet{ye2007svm}.

Due to the higher parallelization potential of the \gls{lssvm}, we decided to base our \gls{plssvm} library on the least squares formulation of \citet{Suykens1999}.
Our goal is to provide a \gls{hpc} implementation of a \gls{lssvm} suitable for modern massively parallel heterogeneous hardware.

Do et al.~\cite{do2009lssvm, do2008novel} developed an \gls{lssvm} implementation using \gls{cuda} to support NVIDIA \glspl{gpu}.
However, they only investigated their implementation on data sets with less than $200$ dimensions. 
Our \gls{plssvm} implementation can easily cope with data sets having more than $\num{65000}$ data points with more than $\num{16000}$ features, as shown in \autoref{sec:result}.
Other \gls{lssvm} implementations were developed, e.g., an \gls{lssvm} purely written in Python using PyTorch by \citet{drumond2021lssvm}.

In this paper, we describe our \texttt{C++17} \gls{plssvm} library, which efficiently brings \glspl{svm} to massively parallel accelerators, and we present and analyze some of our first performance results.
We have implemented the basic functionality of the most widely used \gls{svm} library, LIBSVM~\cite{CC01a}, as a drop-in alternative with significant acceleration aided by \glspl{gpu}.
However, sparse data sets, where all but a few feature entries are zero, are treated as if they would represent dense data, i.e., explicitly representing zeros where necessary.
As of now, our implementation only supports binary classification.

Our aim is to provide vendor-independent scalable performance for high-end compute nodes.
To the best of our knowledge, our \gls{plssvm} library is the first \gls{svm} implementation, let alone \gls{lssvm} implementation, that supports multiple backends, in particular OpenMP, \gls{cuda}, \gls{opencl}, and SYCL (hipSYCL and DPC++).
This allows us to support a broad spectrum of hardware, e.g., \glspl{gpu} from different vendors like NVIDIA, AMD, and Intel.
This distinguishes our approach from most previous implementations, which are restricted to NVIDIA \glspl{gpu} due to their focus on \gls{cuda}.
A thorough comparison of the differences and advantages of the individual languages and frameworks for \glspl{svm} will be covered in a future work.
Furthermore, we support multi-\gls{gpu} execution for the linear kernel.

This paper is structured as follows: In \autoref{sec:theory} we discuss the basics of \glspl{svm} and the transformation into their \gls{lssvm} representation, suitable for highly parallel processing.
In \autoref{sec:impl}, we then deduce important implementational details from the \gls{lssvm} equations.
In particular, we tackle the problem by solving the resulting equation with the \gls{cg}~\cite{hestenes1952methods} algorithm and a parallel matrix assembly.
In \autoref{sec:result}, we examine our achieved speedup factors, while maintaining accuracies on par with the \gls{smo} approaches: up to $10$ for LIBSVM on the \gls{cpu} and up to $14$ compared to ThunderSVM on a single \gls{gpu}.
Additionally, we investigate the runtime characteristics of the individual components of our library as well as the influence of epsilon (the relative residual used in the \gls{cg} algorithm) on the runtime and accuracy, and finally we demonstrate that our approach is scalable with a parallel speedup of up to $\num{74.7}$ on a many-core \gls{cpu} with $256$ threads and on four NVIDIA A100 \glspl{gpu} with a parallel speedup of $\num{3.71}$.
\Autoref{sec:conclusion} concludes this paper with a summary and an outlook on future work.


\section{Support Vector Machines}
\label{sec:theory}

In this section, we describe \glspl{svm} and their transformation into a least squares problem.
The transformation to the \gls{lssvm} is crucial as the state-of-the-art \gls{smo} approach is inherently sequential and, therefore, unsuitable for achieving high speedups on massively parallel devices.


\subsection{Training}
\label{sec:training}

To learn an \gls{svm} classification model a $d$-dimensional data set $T$ with $m$ data points $\vec{x}_i$ and their corresponding labels $y_i$ is required for training.
Based on $T$, an \gls{svm} learns to distinguish between different classes.
We assume a binary classification problem and, w.l.o.g., that classes are labeled $y_i=\pm1$.
The actual learning process is to find the hyperplane $\langle \vec{w},\vec{x}_i\rangle + b = 0$ that divides the two classes best.
Therefore, the normal vector $\vec{w}$ is scaled in a way that the margin, and thus the width of the separation, is $\frac{2}{\Vert \vec{w} \Vert}$.
The bias $b$ indicates, with $\frac{b}{\Vert \vec{w} \Vert}$, the distance of the hyperplane to the origin.
This can be expressed by the following inequality:
\begin{equation}
    \label{fnc:nebenbedingung}
    y_i \cdot (\langle \vec{w},\vec{x}_i\rangle + b) \geq 1 \qquad\forall i.
\end{equation}
If the data cannot be separated linearly, this inequality cannot be solved.
In order to train such problems as well, a vector $\vec{\xi}$ of positive slip variables $\xi_i$ is introduced:
\begin{equation}
    \label{fnc:nebenbedingung_schlupf}
    y_i \cdot (\langle \vec{w},\vec{x}_i\rangle + b) \geq 1 -\xi_i \qquad \forall i.
\end{equation}
At this point the goal is to find the hyperplane with the largest bandwidth, where both, the errors and the sum of the violations by $\vec{x}_i$, should be minimized.
In order to weight the two terms to minimize, a positive weighting constant $C$ is introduced.
Now we can describe the problem mathematically by:
\begin{equation}
    \label{fnc:nonlinneben}
    \begin{split}
        &\min_{\vec{w},b,\vec{\xi}}\left(\frac{1}{2}\Vert  \vec{w}\Vert_2^2 + C\sum_{\forall i} \xi_i\right)\quad\hspace*{.4em}\mid C > 0,\\
        \text{s.t.:} &\quad y_i \cdot (\langle \vec{w},\vec{x}_i\rangle + b)  \geq 1 - \xi_i \quad\mid \xi_i\geq0 \qquad\forall i.\\
    \end{split}
\end{equation}


\subsection{Classification}
\label{sec:classification}

To classify a data point, the \gls{svm} computes on which side of the previously learned hyperplane the data point lies. 
This requires the normal vector $\vec{w}$ and the bias $b$ to compute:
\begin{equation}
    \label{fnc:entscheidungsfunktion}
    \hat{y} = \sgn\left(\langle \vec{w},\vec{\hat{x}}\rangle + b\right).
\end{equation}
The sign of $\hat{y}$ determines the class the test point $\vec{\hat{x}}$ is more likely to belong to.
The distance from the hyperplane indicates the confidence in a correct classification.


\subsection{Least Squares Support Vector Machines}
\label{sec:lssvm}

In its least squares form, the secondary condition in \autoref{fnc:nonlinneben} is no longer understood as an inequality but as an equality.
In contrast to the classic \gls{svm}, not only a few data points are rated for classification, but the \gls{lssvm} determines a weighting proportional to the distance between each data point and the hyperplane.
Therefore, in the \gls{lssvm} all data points are interpreted as support vectors.
In this case, the weighting can also be negative in contrast to the classical \gls{svm}.
Because of this, the $\xi_i$ are squared so that only positive distances are added. 
To compensate for that, the weighting $C$ is halved:
\begin{equation}
    \label{fnc:lssvn}
    \begin{split}
        &\min_{\vec{w},b,\vec{\xi}}\left(\frac{1}{2}\Vert  \vec{w}\Vert_2^2 + \frac{C}{2}\sum_{\forall i} \xi_i^2\right),\\
        \text{s.t.:} &\quad y_i \cdot (\langle \vec{w},\vec{x}_i\rangle + b)  = 1 - \xi_i \quad\mid \xi_i\geq0 \qquad\forall i.\\
    \end{split}
\end{equation}
For a detailed comparison between the \gls{svm} and the \gls{lssvm} regarding their models and accuracies, see \citet{ye2007svm}.


\subsection{Dual Problem}
\label{sec:dual_problem}

The hyperplane determined by the (LS-)\gls{svm} is dependent on---and thus clearly determined by---the support vectors.
Therefore, the normal vector $\vec{w}$ can also be set up as a linear combination of these support vectors:
\begin{equation}
    \vec{w}=\sum_{i=1}^{m}\alpha_i y_i \vec{x}_i \ \quad\mid \alpha_i \in \mathbb{R}.
\end{equation}
While $\alpha_i\in \vec{\alpha}$ is equal to zero for all $\vec{x}_i$ that are not support vectors and, therefore, not weighted, $\alpha_i$ is non-zero for the support vectors.
For the \glspl{lssvm} normally all $\alpha_i$ are non-zero.
Using the Lagrange multipliers $\alpha_i$ and the Karush-Kuhn-Tucker conditions, the minimization problem can be reformulated into its dual form, as proposed by~\citet{saunders1998ridge}.
On this basis, \citet{Chu2005} formulated, under elimination of $\vec{\xi}$ and $\vec{w}$, the following maximization problem:
\begin{equation}
    \label{func:kruscal}
    \begin{split}
        \max_{\vec{\alpha}} & \left(\sum_{i=1}^{m}\alpha_i - \frac{1}{2} \sum_{i=1}^{m}\sum_{j=1}^{m}\alpha_i\alpha_jy_iy_j\langle \vec{x}_i,\vec{x}_j\rangle\right),\\
        \text{s.t.:}  & \quad 0\leq  \alpha_i \leq C, \hspace*{1em}\sum_{i=1}^{m}\alpha_iy_i = 0.
    \end{split}
\end{equation}


\subsection{Kernel Trick}
\label{sec:kerneltrick}

In machine learning, the linearity of methods is often exploited, and the so-called kernel trick is used to classify non-linear problems with linear classifiers.
A possible kernel function is a mapping $k : X \times X \rightarrow \mathbb{R}$ given the input space $X \in \mathbb{R}^d$, if it is defined on an inner product space $(F, \langle\cdot , \cdot\rangle)$ and, if a feature mapping $\varPhi : X \rightarrow F$ to the feature space $F \in \mathbb{R}^{\hat{d}}$ with $\hat{d} \geq d$ exists \cite{Hofmann2008}:
\begin{equation}
	k(\vec{x}_i, \vec{x}_j) = \langle\varPhi(\vec{x}_i), \varPhi(\vec{x}_j)\rangle.
\end{equation}
The scalar product in \autoref{func:kruscal} is now replaced by the kernel function $k$,
\begin{equation}
	\label{func:kernel}
	\begin{split}
		\max_{\vec{\alpha}} & \left(\sum_{i=1}^{m}\alpha_i - \frac{1}{2} \sum_{i=1}^{m}\sum_{j=1}^{m}\alpha_i\alpha_jy_iy_j \cdot k(\vec{x}_i,\vec{x}_j)\right),\\
		\text{s.t.:}  & \quad 0\leq  \alpha_i \leq C, \hspace*{1em}\sum_{i=1}^{m}\alpha_iy_i = 0.
	\end{split}
\end{equation}
The classification function (\autoref{fnc:entscheidungsfunktion}) must also be adapted.
This is achieved by using Lagrange multipliers and the substitution of the scalar product with the kernel function,
\begin{equation}
	\begin{split}
		f(\vec{x})&=\sgn\left(\sum_{i = 1}^{m}\alpha_i y_i \cdot k(\vec{x}_i,\vec{x}) + b\right).
	\end{split}
\end{equation}
In our \gls{plssvm} implementation the following three kernel functions $k(\vec{x}_i, \vec{x}_j)$ are available:

\begin{tabular}{llr}
	linear:     & $\langle \vec{x}_i, \vec{x}_j \rangle$                     &                                     \\
	polynomial: & $(\gamma\cdot\langle \vec{x}_i, \vec{x}_j \rangle + r)^d,$ & $ \gamma > 0, d\in\mathbb{Z}$ \\
	radial:     & $\exp(-\gamma\cdot\Vert \vec{x}_i - \vec{x}_j \Vert^2_2),$   & $ \gamma > 0$                       \\
\end{tabular}


\subsection{System of Linear Equations}
\label{sec:lgs}

Following \citet{Suykens1999}, \autoref{func:kernel} can be transformed into a system of linear equations,
\begin{equation}
	\label{fnc:equation}
	\begin{bmatrix}
		Q           & \vec{1}_m \\
		\vec{1}^T_m & 0
	\end{bmatrix}
	\cdot
	\begin{bmatrix}
		\vec{\alpha} \\
		b
	\end{bmatrix}
	=
	\begin{bmatrix}
		\vec{y} \\
		0
	\end{bmatrix},
\end{equation}
where $Q \in \mathbb{R}^{m\times m}$ is a quadratic matrix, $\vec{1}_m$ is a vector with $m$ times the number one, and $Q_{ij} = k(\vec{x}_i, \vec{x}_j) + \frac{1}{C}\delta_{ij}$.
The Kronecker Delta $\delta_{ij}$ is equal to zero everywhere, except for $i = j$.
To reduce the matrix size, \citet{Chu2005} divided the matrix $Q$ into four parts: the lower row $\vec{q}^T$ and the right column $\vec{q}$, the last entry of the matrix $Q_{m,m}$ and $\bar{Q}$, the remaining part of $Q$,
\begin{equation}
	Q = \begin{bmatrix}
		\bar{Q} & \vec{q} \\ \vec{q}^T &Q_{m,m}
	\end{bmatrix}.
\end{equation}
With this partitioning, a new matrix $\tilde{Q}$ is constructed:
\begin{equation}
	\label{func:lgs}
	\begin{split}
		\tilde{Q} :=& \bar{Q}-\vec{1}_{m-1}\cdot \vec{q}^T- \vec{q} \cdot \vec{1}^T_{m-1} + Q_{m,m} \cdot \vec{1}_{m-1}\cdot \vec{1}^T_{m-1}.
	\end{split}
\end{equation}
\Autoref{fnc:equation} can now be represented in a smaller system \cite{Chu2005}:
\begin{equation}
	\label{fnc:small_equation}
	\begin{split}
		\tilde{Q}\cdot\vec{\tilde{\alpha}}=\vec{\bar{y}}-y_m\cdot \vec{1}_{m-1},\\
	\end{split}
\end{equation}
where $\vec{\tilde{\alpha}}$ is unknown and $\vec{\bar{y}}$ is defined as the first $m - 1$ values of $\vec{y}$.
From \autoref{fnc:small_equation}, the bias $b$ and the weights $\vec{w}$ can be derived by:
\begin{equation}
    \begin{array}{ll}
    	b &= y_m + Q_{m,m} \cdot \langle\vec{1}_{m-1}, \vec{\tilde{\alpha}}\rangle - \langle\vec{q}, \vec{\tilde{\alpha}}\rangle \\
    	\vec{w} &= \sum\limits_{i=1}^{m}\alpha_i\varPhi(\vec{x}_i).
	\end{array}
\end{equation}


\subsection{Parallelizability}
\label{sec:parallel}

From an algorithmic point of view, the \gls{smo} method takes one point from each class and computes a hyperplane between them.
Then another pair is drawn, seen if the point pair is in the current margin, and if so, the existing hyperplane is adjusted accordingly.
This procedure is repeated until the next hyperplane adjustment would change it less than some $\epsilon > 0$.

It is evident that this algorithm is inherently sequential and cannot be parallelized without special adaption.
Therefore, optimized \gls{smo} implementations do not use point pairs but point groups. 
They parallelize the hyperplane adjustment within these groups, whereby the computation becomes more complex the larger these groups grow.
Thus, these \gls{smo} variations have parallelization potential but are not suitable for massively parallel \gls{hpc} hardware.
Additionally, the larger the groups grow, the more the classification behavior corresponds to the \gls{lssvm} behavior.

In contrast, solving the \gls{lssvm} problem can be translated to solving a system of linear equations.  This matrix is positive, symmetric definite, for which there are well-known and efficient algorithms suitable for massively parallel hardware.


\section{Implementation}
\label{sec:impl}

We developed the \gls{plssvm} library for maximum performance and portability.
Therefore, we have not only implemented the \gls{lssvm} using a single framework but with multiple ones: we cover OpenMP, \gls{cuda}, \gls{opencl}, and SYCL (hipSYCL and DPC++).
This opens up the possibility to target a broad variety of different hardware architectures, in particular \glspl{cpu} and \glspl{gpu} from NVIDIA, AMD, and Intel.
All backends are optional, i.e., they are only conditionally included if the necessary hardware and software stack is available. 
The actual used backend can be selected at runtime.

We use \texttt{C++17} focusing on modern best practices and support switching between double and single precision floating point types by changing a single template parameter.
The training of \gls{plssvm} can be split into four steps: (1) read the whole training data set used to set up the system of linear equations (\autoref{fnc:small_equation}), (2) load the training data onto the device, (3) solve the system of linear equations and write-back the solution to the host memory, and (4) save the learned support vectors with the corresponding weights to a model file.

Currently, our implementation only supports dense data for calculations.
If sparse data sets are used, they are at first converted into a dense representation by filling in zeros.

In the following, we investigate implementational details of these four steps.
Hereby, the main focus lies on the solution of the system of linear equations, since this is the most computationally intensive part.

\subsection{Data Layout}
\label{sec:data_layout}

While the data points are initially read into an irregular $2D$ data structure, they are later transformed into a $1D$ vector laid out consecutively in memory.
Since the data is accessed dimension-wise during execution, the data vector is not created arranging the individual points in row-major order, but it is constructed in column-major order.
This has the advantage that it is significantly more cache-efficient for our indexing scheme on \glspl{gpu} and thus results in better performance.


\subsection{Solving the System of Linear Equations}
\label{sec:solve_lgs}

To solve the system of linear equations (\autoref{fnc:small_equation}), we use the \gls{cg} method.
This is possible because the matrix $\tilde{Q}$ is symmetric and positive definite \cite{Chu2005}.
Hereby, we implement a variant of \citet{shewchuk1994introduction}.

Since $\tilde{Q}$ has $(m - 1)^2$ entries, it is not feasible for large training data sets to store the matrix completely in memory.
Therefore, the matrix is only represented implicitly: each entry $\tilde{Q}_{i,j}$ is recalculated for each use according to \autoref{func:lgs}:
\begin{equation}
	\label{func:Qij}
	\arraycolsep=1.4pt
	\begin{array}{ll}
		\tilde{Q}_{i,j} &= \mathrm{k}(\vec{x}_i,\vec{x}_j) + \dfrac{1}{C} \cdot \delta_{i,j} - \mathrm{k}(\vec{x}_m,\vec{x}_j) \\
		&- \mathrm{k}(\vec{x}_i,\vec{x}_m) + \mathrm{k}(\vec{x}_m, \vec{x}_m) + \dfrac{1}{C}.
	\end{array}
\end{equation}


\subsection{Optimizations}
\label{sec:optimizations}

Matrix-vector multiplications are the most computational expensive task in the \gls{cg} algorithm.
Therefore, we decided to calculate this function in parallel on the available \glspl{gpu}, carefully tuning the implementation using multiple optimization strategies to achieve the best performance possible.
Note that the \gls{cpu} only OpenMP implementation is currently not as well optimized as the \gls{gpu} implementations used by the other backends.

The following sections exclusively use the \gls{cuda} notation for the memory and execution models.\footnote{For a mapping between the \gls{cuda} notations and the corresponding \gls{opencl} and SYCL names see \url{https://developer.codeplay.com/products/computecpp/ce/guides/sycl-for-cuda-developers/migration} (2022-01-20).}
However, all optimizations were also applied to the \gls{opencl} and SYCL backends.

\subsubsection{Blocking}
\label{sec:blocking}

The calculations are divided into individual blocks, which in turn are calculated independently in parallel.
To ensure that as few boundary conditions as possible have to be checked (i.e., avoiding branch divergence), we decided to introduce a padding that is always at least the size of a full block.
Since the matrix is symmetric, the effort can be halved by computing only the upper triangular matrix.
The omitted entries can then be mirrored.
In our approach, we simply spawn threads for the whole matrix, but only the threads corresponding to blocks above the diagonal are used for the actual calculations.
Thus, each thread with an index $i \geq j$ computes $\tilde{Q}_{i,j}$. 
This technique is inexpensive since thread creation on \glspl{gpu} is rather lightweight.

\subsubsection{Caching}
\label{sec:caching}

Since two vector-entries of $\vec{q}$ are required for every calculation of $\tilde{Q}_{i,j}$, it is worthwhile to precalculate $\vec{q}$.
This reduces the amount of expensive scalar products per matrix element from three to one by storing $m - 1$ values.

\subsubsection{Block-Level Caching}
\label{sec:block_level_caching}

\glspl{gpu} have different memories at their disposal.
The very fast registers are separated for each thread, as opposed to the slower shared memory, which is shared by all threads in a thread block.
However, shared memory is still much faster than global memory accessible by all threads, even across thread blocks.

Therefore, it is necessary to employ some form of blocking to utilize the faster shared memory.
For each thread block, $blocksize\cdot 2$ many data points are needed to calculate $blocksize^2$ many kernel operations (e.g., scalar products for the linear kernel; see \autoref{sec:kerneltrick}).
We start by loading a few features of the data points, which are needed for the first operations, into the shared memory.
To utilize the full bandwidth of the shared memory, which is much higher compared to the bandwidth of the global memory, it becomes important to evenly distribute the memory reads over all warps.
Next, we proceed with the actual calculations, followed by loading the next features of the data points into the shared memory, and so on.

\subsubsection{Thread-Level Caching}
\label{sec:thread_level_caching}

While our block-level caching operates between the shared memory and the global memory, we also apply blocking inside each thread of a thread block between its registers and the shared memory.

Our implementation allows both blocking sizes to be changed during compilation.
This way, the optimizations can be adapted to the given hardware to maximize the performance gains.

\subsubsection{Distribute Blocks Across Multiple GPUs}
\label{sec:multi_gpu}

Our implementation is capable of utilizing multiple \glspl{gpu} for the linear kernel.
In order to achieve an even utilization of all \glspl{gpu}, we do not divide the data set itself, but we split all data points feature-wise.
For example, if the data is in a ten-dimensional space (i.e., each data point has ten features) and two \glspl{gpu} are available, each data point is split into two five-dimensional data points and each \gls{gpu} is assigned one of these.
Due to the linearity of the linear kernel, only the result vectors of the single devices have to be summed up.
The polynomial and radial kernels do not currently support multi-\gls{gpu} execution. 
The remaining calculations stay the same, no matter how many devices are used.
This approach also reduces the memory usage per \gls{gpu}, so that larger data sets can be learned.


\section{Results}
\label{sec:result}

This chapter presents the experimental results for our \gls{lssvm} implementation.
At first, we compare our \gls{plssvm} (\href{https://github.com/SC-SGS/PLSSVM/tree/v1.0.1}{v1.0.1}) with the \gls{cpu} versions of the normal and dense\footnote{\url{https://www.csie.ntu.edu.tw/\~cjlin/libsvmtools/\#libsvm\_for\_dense\_data} (2022-01-20)} implementations of LIBSVM (3.25) and ThunderSVM (git commit \href{https://github.com/Xtra-Computing/thundersvm/tree/55ee783952d1c95118314c256a844183eef3bbe7}{55ee783}) as well as with the \gls{gpu} implementation of ThunderSVM.
We do not include the \gls{cuda} implementation of LIBSVM in our tests, since the normal training workflow does not support \gls{gpu} execution, only the cross validation is implemented for a \gls{gpu}. 
Afterward, we investigate the runtime behavior of the different components of \gls{plssvm}.
We also discuss the influence of the epsilon parameter (the relative residual of the \gls{cg} method) on the resulting runtime and accuracy.
Finally, we show that our implementation indeed scales on many-core \glspl{cpu} and multiple \glspl{gpu}.


\subsection{Hardware Platforms}
\label{sec:hardware}

We used two different hardware platforms for our performance measurements.
All the \gls{cpu} runtimes were measured on a machine with two $64$ cores / $128$ threads AMD EPYC 7742 \glspl{cpu} @$\SI{2.25}{\giga\hertz}$ and $\SI{2}{\tera\byte}$ DDR4 RAM.
The runtimes involving \glspl{gpu} were computed on a machine with four NVIDIA A100 \glspl{gpu} (each: $6912$ shader units @$\SI{1.095}{\giga\hertz}$, $\SI{40}{\giga\byte}$ HBM2 memory, $\SI{1.555}{\giga\byte\per\second}$ memory bandwidth, and $\SI{9.7}{\tera\flops}$ (FP64) peak performance), two $64$ cores / $128$ threads AMD EPYC 7763 \glspl{cpu} @$\SI{2.45}{\giga\hertz}$, and $\SI{1}{\tera\byte}$ DDR4 RAM. 
However, on the \gls{gpu} machine we ensured that only $64$ physical cores, all on the same socket, were used.
The \gls{gpu} machine is equipped with NVLink 3.0 which is, however, not used by our library since we have no direct \gls{gpu}-to-\gls{gpu} communication.
All measurements were performed with double precision floating point types (FP64).


\subsection{Data Sets and Experimental Setup}
\label{sec:data_sets_and_exp_setup}

In order to report scaling runtimes with respect to a growing number of data points and features, we decided to focus on synthetically generated dense data sets.
The training files are created by Python's scikit-learn single-label generator \texttt{make\_classification} \cite{scikit-learn}.
These synthetic data sets can be reproduced with the \texttt{generate\_data.py} script located in the \gls{plssvm} GitHub\footnote{\url{https://github.com/SC-SGS/PLSSVM/blob/v1.0.1/utility_scripts/generate_data.py} (2022-01-20)} using the \texttt{problem} type \enquote{planes}.
We have generated separate new data files for each individual run to average over a variety of different spatial arrangements of the same problem.

The two generated clusters are adjacent to each other and overlap with a low probability in a few points.
Additionally, one percent of the labels were set randomly to ensure some noise in the generated data.
The \gls{smo} approach would be a better fit for well separable data.
However, real-world data is rarely free of noise and in the rarest cases also well and clearly separable.
The more noisy and ambiguous the training data is, the better the \gls{lssvm} is suited (for the theoretical background, see \cite{ye2007svm}).
Therefore, the data for this analysis were selected with a little noise but still relatively well separable.
This ensures that the data suits the \gls{smo} approach while representing the structure of real-world-like data.
The number of data points and features for all synthetically generated data sets are power of twos.
However, our library is not limited or specifically optimized to sizes of power of twos, it is rather used for convenience in the  logarithmic plots.

In addition to these synthetically generated data sets, we also used the real-world SAT-6 Airbone data set \cite{Basu2015}. 
This data set consists of images displaying six different land cover classes. 
Since we currently only support binary classification, we mapped the labels of all man-made structures (buildings and roads) to $-1$ and the labels of the remaining classes (barren land, trees, grassland, and water bodies) to $1$.
The data set is split into training data with $\num{324000}$ ($\num{193729}$ with label $-1$ and $\num{130271}$ with label $1$) images of size $28 \times 28$ with $4$ color channels (RGB-IR) resulting in $\num{3136}$ features per image and test data with $\num{81000}$ images.
All features are scaled to values between $[-1, 1]$ using LIBSVM's \texttt{svm-scale}.
\todo{anschauen Review 1 Separierbarkeit? $\rightarrow$ Dirk fragen}

Since the methods have different termination criteria, it is not trivial to compare the runtimes with each other.
We compare the runtimes by adjusting the epsilon of the loss function for the \gls{smo} methods and the epsilon of the relative residual used in the \gls{lssvm} algorithm. 
We start with $\num{0.1}$ and increment the epsilon in steps of $\times0.1$ (i.e., $\num{0.01}$, $\num{0.001}$, etc.) until an accuracy of more than $\SI{97}{\percent}$ was reached on the training data.
If the training data was non-separable, i.e., we were not able to reach a minimum accuracy of $\SI{97}{\percent}$, we compared the runs that converged in accuracy in the first three digits.
Except for the epsilon, the default values of the libraries were retained.
This implies that no special optimizations  with regard to the used hardware were performed for any of the three used \gls{svm} implementations.

All runtimes were averaged over at least ten measurement runs.
Our comparison is limited to NVIDIA \glspl{gpu}, since ThunderSVM is only implemented in CUDA.
To make the comparison fairer, we use our \gls{plssvm} \gls{cuda} backend for the \gls{gpu} runs and the OpenMP backend for the \gls{cpu}.
However, this does not limit the results to the \gls{cuda} backend but generally holds for all our \gls{gpu} backends, since they all have similar runtime behaviors with respect to the number of data points and features.

\begin{table}
    \renewcommand{\arraystretch}{1.3}
    \caption{Runtime examples for the different backends on different \glspl{gpu} for $2^{15}$ data points with $2^{12}$ features each reaching approx. $\SI{93.76}{\percent}$ accuracy.}
    \label{table:backends}
    \centering
    \begin{tabular}{crrr}
           hardware     &        CUDA         &       OpenCL        &   SYCL    \\\hline\\[-.9em]
        NVIDIA GTX 1080 Ti   & $\SI{369.57}{\second}$ & $\SI{380.98}{\second}$ & $\SI{738.46}{\second}$ \\
        NVIDIA RTX 3080      & $\SI{251.66}{\second}$ & $\SI{266.00}{\second}$ & $\SI{269.96}{\second}$ \\
        NVIDIA P100    & $\SI{92.87}{\second}$ & $\SI{97.85}{\second}$ & $\SI{329.06}{\second}$ \\ 
        NVIDIA V100    & $\SI{37.96}{\second}$ & $\SI{55.48}{\second}$ & $\SI{72.13}{\second}$ \\
        AMD Radeon VII & --- & $\SI{152.05}{\second}$ & $\SI{189.21}{\second}$ \\
        Intel UHD Graphics Gen9 P630 & --- & $\SI{3788.43}{\second}$ & $\SI{7355.93}{\second}$ \\\hline
    \end{tabular}
	\vspace*{-1em}
\end{table}

\autoref{table:backends} displays some example runtimes for our \gls{cuda}, \gls{opencl}, and SYCL (DPC++ for the Intel \gls{gpu}, hipSYCL otherwise) backends on multiple NVIDIA \glspl{gpu} as well as an AMD and Intel \gls{gpu}.
The runtimes only include the actual learning part on the \glspl{gpu} to factor out the influences of the different \glspl{cpu} in the different systems.
On the NVIDIA \glspl{gpu}, the \gls{cuda} backend is the fastest, closely followed by \gls{opencl}.
hipSYCL is only slightly slower than \gls{opencl} for compute capabilities of $\num{7.0}$ or newer.
However, for \glspl{gpu} with an older compute capability, hipSYCL is over three times slower than \gls{cuda} or \gls{opencl} indicating that \gls{plssvm} uses a feature which hipSYCL does not efficiently map to older NVIDIA \glspl{gpu}.
On the AMD \gls{gpu}, hipSYCL is again slightly slower compared to \gls{opencl}.
On the Intel \gls{gpu}, our SYCL backend is two times slower than our \gls{opencl} backend.
More profiling has to be done to study the exact reasons for the poor performance of our SYCL backend in some setups compared to the other backends.
However, this analysis and the influence of the hardware characteristics of the different \gls{gpu} vendors on the runtime goes beyond the scope of this paper.


\subsection{Runtime Comparison}

\begin{figure}
	\centering
	\subfloat[\gls{cpu} runtimes relative to the number of data points with $2^{10}$ features.\label{fig:runtime:cpu:datapoints}]{\includegraphics[width=.49\linewidth]{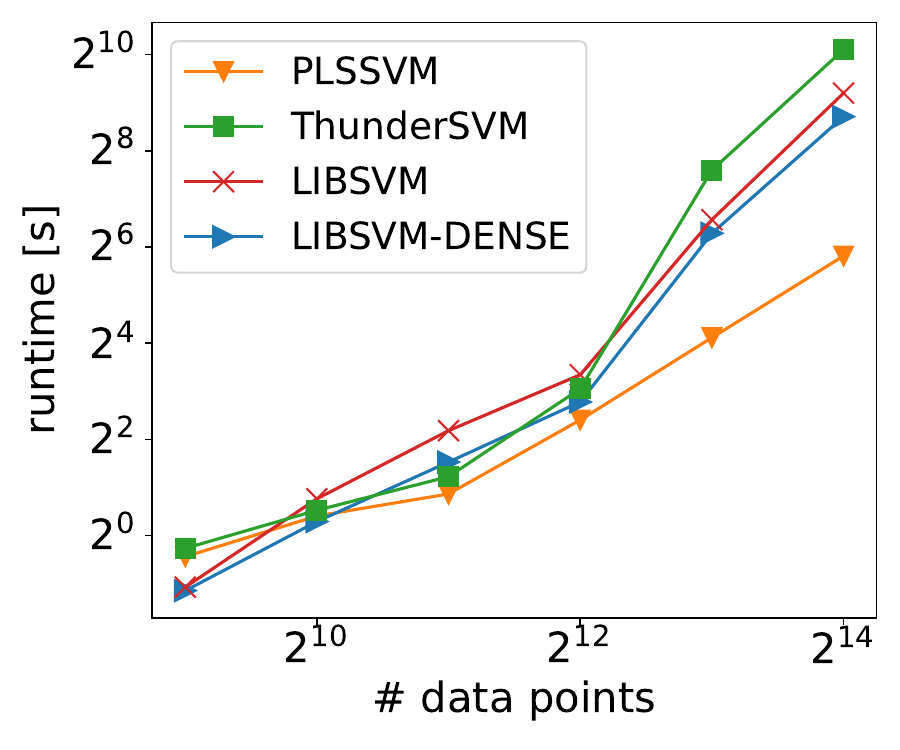}}
	\hfill
	\subfloat[\gls{cpu} runtimes relative to the number of features with $2^{13}$ data points.\label{fig:runtime:cpu:features}]{\includegraphics[width=.49\linewidth]{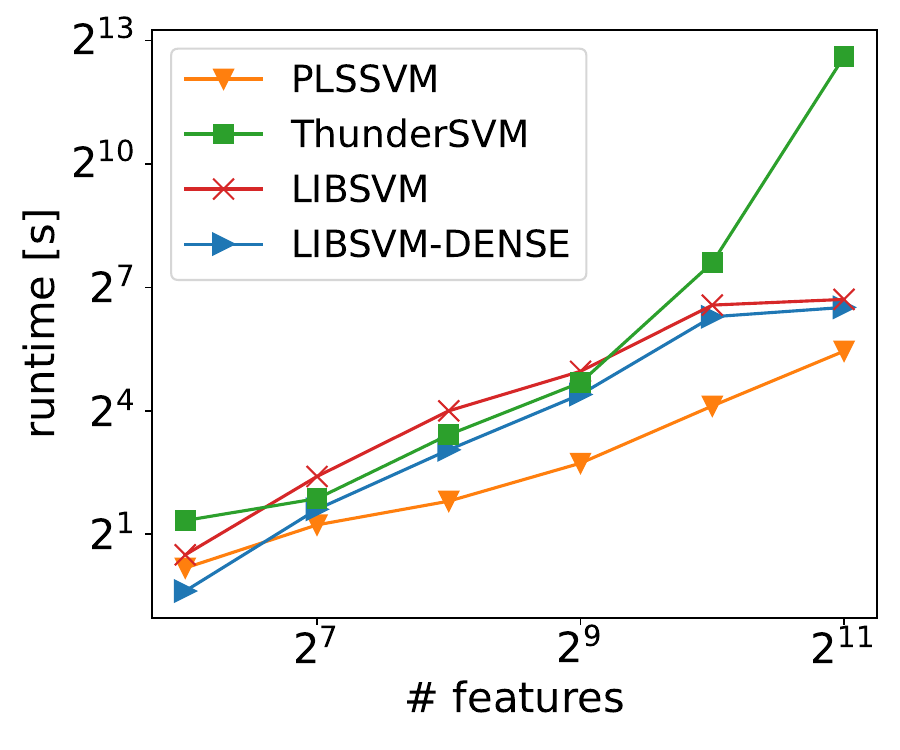}}
	\\
	\subfloat[\gls{gpu} runtimes relative to the number of data points with $2^{12}$ features.\label{fig:runtime:gpu:datapoints}]{\includegraphics[width=.49\linewidth]{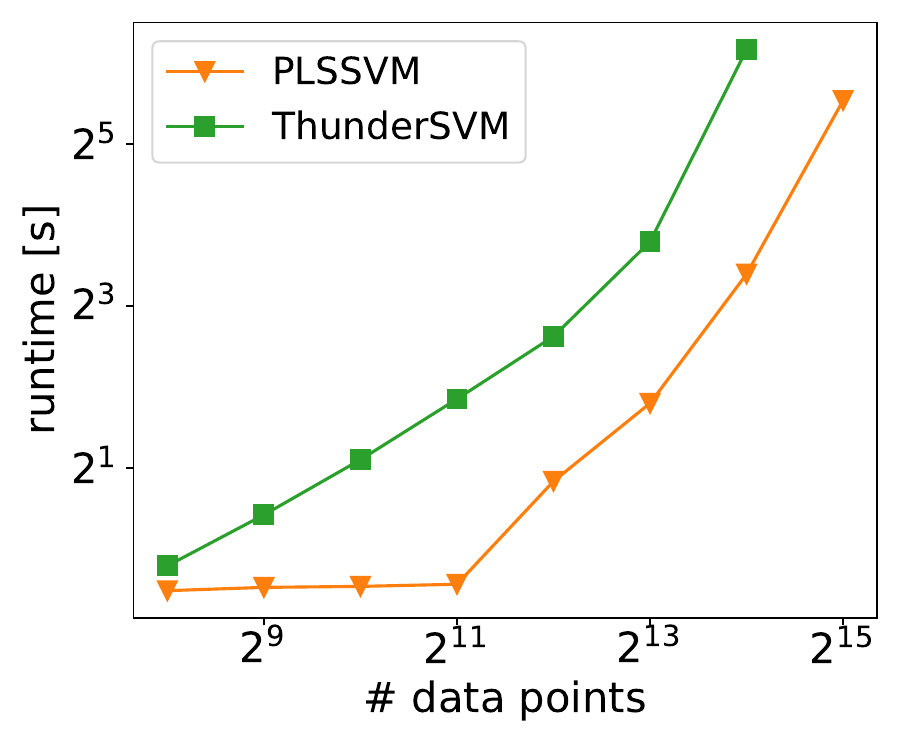}}
	\hfill
	\subfloat[\gls{gpu} runtimes relative to the number of features with $2^{15}$ data points.\label{fig:runtime:gpu:features}]{\includegraphics[width=.49\linewidth]{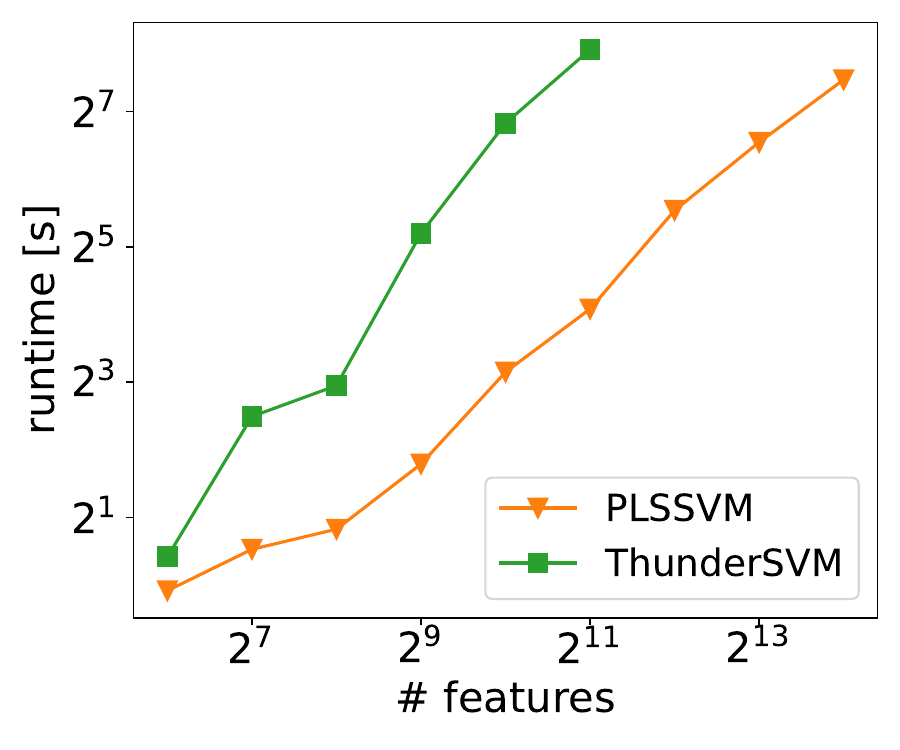}}
	\caption{Double logarithmic runtime comparison for sparse/dense LIBSVM, ThunderSVM, and \gls{plssvm} on both \gls{cpu} and \gls{gpu} in relation to the number of data points and features. 
	The runtimes were measured with an epsilon, such that the resulting models reach approximately $\SI{97}{\percent}$ accuracy on the training data or after convergence of the accuracy in the first three decimal places.}
	\label{fig:runtime:compare}
	\vspace*{-1em}
\end{figure}

First, we discuss the overall runtime of our implementation using the \gls{cpu} and \gls{gpu} for training the \gls{svm}; see \autoref{fig:runtime:compare}.
We only consider the linear kernel for the synthetic data sets, since the behavior can be projected one-to-one to the other kernels.

The runtime behavior for \gls{plssvm}, ThunderSVM, and LIBSVM on the \gls{cpu}, for a fixed number of features and a varying number of data points, is shown in \autoref{fig:runtime:cpu:datapoints}.
All three \gls{smo} implementations behave very similarly, with the dense implementation of LIBSVM having a slight advantage over its sparse implementation.
ThunderSVM performs better than LIBSVM in the medium size range, while in the extremes the LIBSVM variants are faster.
However, all \gls{smo} implementations exhibit a steeper slope, compared to the \gls{lssvm} method used in \gls{plssvm}.
We out-scale both LIBSVM variants, starting with $2^{11}$ data points.
For $2^{14}$ data points in $2^{10}$ dimensions we need approximately $\SI{56.2}{\second}$ to train the \gls{plssvm} while LIBSVM already needs $\SI{7.0}{\minute}$ (dense) respectively $\SI{9.8}{\minute}$ (sparse) and ThunderSVM performs worst with $\SI{18.3}{\minute}$.

In \autoref{fig:runtime:cpu:features}, we examine the scalability with respect to the number of features, while the number of data points was fixed to $2^{13}$.
Increasing the number of features, \gls{plssvm} scales (up to $2^{10}$) slightly better than LIBSVM and significantly better than ThunderSVM.
The behavior for the LIBSVM variants changes at $2^{12}$, but this has not been investigated further.

\Autoref{fig:runtime:gpu:datapoints} and \ref{fig:runtime:gpu:features} show the same experimental setup but now using a single \gls{gpu} instead of a \gls{cpu}.
The runtimes for \gls{plssvm} using the \gls{gpu} are orders of magnitude faster than their \gls{cpu} counterpart (see \autoref{fig:runtime:cpu:datapoints}).
\gls{plssvm} needs about $\SI{58}{\second}$ to train a data set with $2^{14}$ data points and $2^{10}$ features on the \gls{cpu}.
Using a single \gls{gpu} reduces this runtime by a factor of $24$ to only $\SI{2.4}{\second}$.
This is surprising since both systems have a comparable computational power with $\SI{9.7}{\tera\flops}$ theoretical peak performance for the NVIDIA A100 \gls{gpu} and with $\SI{4.6}{\tera\flops}$ (base) to $\SI{6.96}{\tera\flops}$ (boost) for the two AMD EPYC 7742 \glspl{cpu}.
This underlines how much optimization potential there still is in our OpenMP implementation.

The average coefficient of variation for the \gls{svm} implementations in \autoref{fig:runtime:cpu:datapoints} and \ref{fig:runtime:cpu:features} are \num{0.26} (PLSSVM), \num{0.92} (ThunderSVM), \num{0.60} (LIBSVM), and \num{0.66} (LIBSVM-DENSE).
This shows that the runtimes of our implementation have drastically less variations between executions when compared to the three \gls{smo} implementations.

\Autoref{fig:runtime:gpu:datapoints} shows the runtimes for $2^{8}$ to $2^{15}$ data points with $2^{12}$ features each.
Up to $2^{11}$ data points, the runtime of \gls{plssvm} does not increase, indicating a significant static overhead using a \gls{gpu}.
However, for large data sets, this overhead is negligible (also see \autoref{fig:components:datapoints}).
The slopes of the \gls{svm} implementations indicate that both have approximately the same overall complexity, with \gls{plssvm} having a drastically smaller constant factor.
Training the \gls{svm} model using a data set with $2^{14}$ data points needs $\SI{10}{\second}$ using \gls{plssvm}. 
ThunderSVM takes $\SI{72}{\second}$, resulting in a runtime increase of a factor of $\num{7.2}$.

The differences become larger if we fix the number of data points to $2^{15}$ and vary the number of features from $2^{6}$ to $2^{14}$.
This time, \gls{plssvm} has a slightly flatter slope than ThunderSVM.
ThunderSVM needs $\SI{241}{\second}$ for training a data set with $2^{11}$ features.
Using \gls{plssvm} this can be reduced by a factor of $\num{14.2}$ to only $\SI{17}{\second}$.
This additionally indicates that our implementation scales better with the number of features compared to ThunderSVM.
Increasing the number of features by a factor of four from $2^{11}$ to $2^{13}$ using \gls{plssvm}, increases the runtime by a factor of roughly $\num{5.5}$.
Actually, we had expected that the factor would be larger, since a four-fold increase of the number of features means an eight-fold increase of the matrix entries.
However, in the selected scenario, the complexity of the problem to learn remains the same. 
It can thus be stated that, despite more features, fewer \gls{cg} iterations are needed to solve the system of linear equations to a similar accuracy.
Note that the observed performance advantage of \gls{plssvm} over ThunderSVM thus depends to some extent on the actual classification problem.

However, it should not become as large as the factor of ThunderSVM in any case, since the complexity of the problem generally does not increase drastically with an increasing number of data points for an \gls{lssvm}.
In this example, the number of required \gls{cg} iterations dropped from an average of $\num{30.5}$ iterations for a data set with $2^{10}$ points and $2^{10}$ features to only $\num{26}$ iterations for $2^{15}$ data points and the same number of features.
Using ThunderSVM the runtime drastically increases by a factor of $\num{239}$ compared to the \gls{cpu} runtimes.

As for the \gls{cpu} runtimes, the average coefficient of variation on the \gls{gpu} is again remarkably smaller for \gls{plssvm} (\num{0.11}) than for ThunderSVM (\num{0.37}).

Looking at profiling results using NVIDIA's Nsight Compute profiling tool, we noticed that ThunderSVM spawns a plethora of small compute kernels on the device (over \num{1600} in our profiled scenario with $2^{14}$ data points and $2^{12}$ features), most of them running significantly less than one millisecond. 
However, the compute kernel with the highest compute intensity only reaches approximately \SI{233}{\giga\flops} which only amounts to \SI{2.4}{\percent} of the A100's theoretical FP64 peak performance.
In contrast, our implementation only spawns \num{3} compute kernels that each have a much higher compute intensity than ThunderSVM's kernels.
The kernel responsible for the implicit matrix-vector multiplication inside the \gls{cg} algorithm reaches over \SI{3.1}{\tera\flops} using double precision resulting in \SI{32}{\percent} FP64 peak performance.

In summary, we can note that our implementation performs comparable to the \gls{smo} implementations on the \gls{cpu} and heavily outperforms ThunderSVM on the \gls{gpu}.

\subsection{The SAT-6 Airbone Real-World Data Set}
\label{sec:real_world_data}

Using the real-world data set, we observe a similar behavior.
On the SAT-6 Airbone training data set, we achieved the highest accuracy with the radial basis function kernel.
Training the \gls{svm} classifier using a single \gls{gpu} takes $\SI{23.5}{\minute}$ for our \gls{plssvm} implementation, resulting in an accuracy of $\SI{95}{\percent}$ on the test data set. 
ThunderSVM needs $\SI{40.6}{\minute}$ for $\SI{94}{\percent}$ accuracy and is thus a factor of $\num{1.73}$ slower than \gls{plssvm}.

\subsection{Runtime Analysis of the PLSSVM Components}
\label{sec:components}

\begin{figure}
	\centering
	\subfloat[\gls{gpu} runtime behavior of the individual \gls{plssvm} components depending on the number of data points with $2^{12}$ features.
	\label{fig:components:datapoints}]{\includegraphics[width=.49\linewidth]{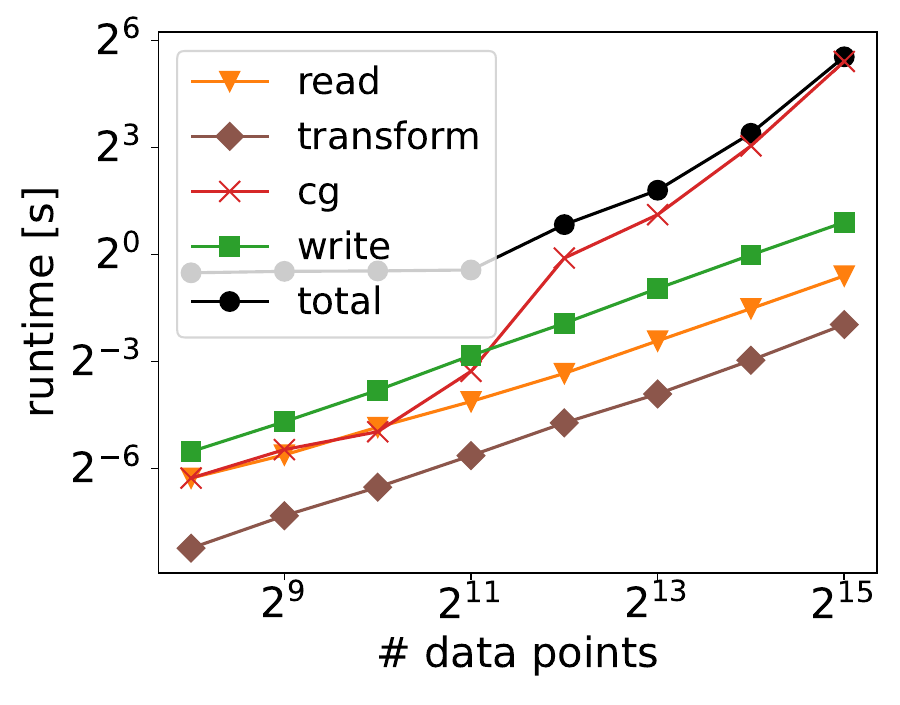}}
	\hfill
	\subfloat[\gls{gpu} runtime behavior of the individual \gls{plssvm} components depending on the number of features with $2^{15}$ data points.
	\label{fig:components:features}]{\includegraphics[width=.49\linewidth]{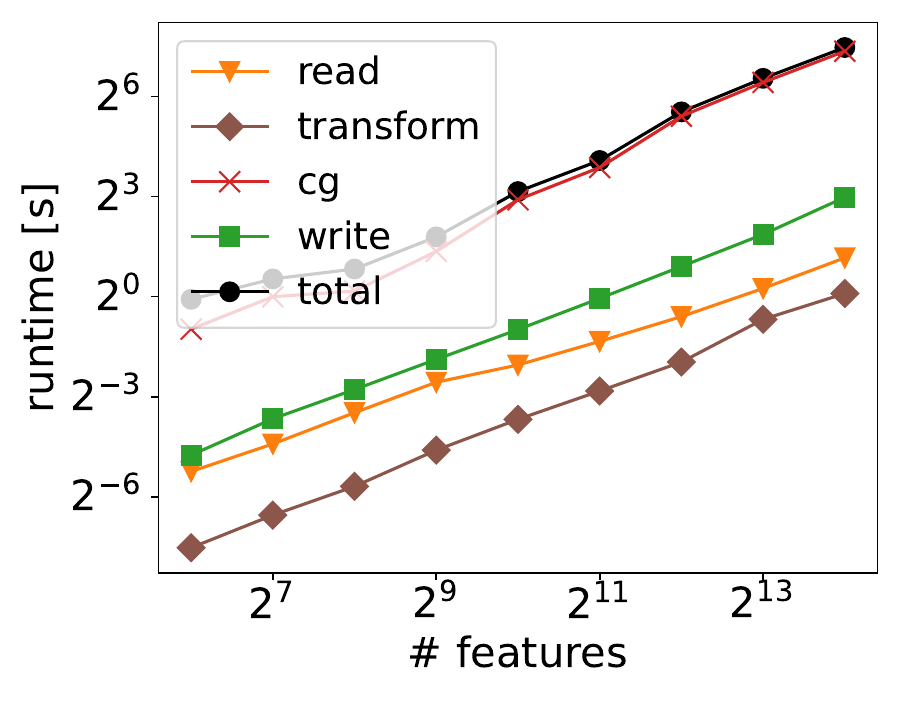}}
	\caption{Double logarithmic runtime scaling behavior of the individual \gls{plssvm} components on a single \gls{gpu}.
	The total runtime depends on the number of \gls{cg} iterations, which are chosen so that the model reaches approximately $\SI{97}{\percent}$ accuracy on the training data or after convergence of the accuracy in the first three decimal places.
	}
	\label{fig:components}
	\vspace*{-1em}
\end{figure}

\Autoref{fig:components} breaks down the runtime of our \gls{plssvm} library using a single \gls{gpu} into its various components:
\begin{itemize}
	\item \textbf{read}: Reads the input training data file and parses its content into a dense $2D$ representation.
	\item \textbf{transform}: Transforms the previously created $2D$ training data structure into a $1D$ vector in \gls{soa} layout for a better caching efficiency.
	This is only relevant for the \gls{gpu} implementations. 
	\item \textbf{cg}: In the \gls{cg} component, the actual system of linear equations is solved using the selected backend.
	\item \textbf{write}: Creates and writes the resulting model file to disk.
	\item \textbf{total}: This represents the runtime of a complete training run including \enquote{read}, \enquote{transform}, \enquote{cg}, \enquote{write} and remaining parts like initializing the backend and hardware.
\end{itemize}

\Autoref{fig:components:datapoints} shows the components' runtimes for data set sizes from $2^{8}$ up to $2^{15}$ using $2^{12}$ features each.
It can be seen that the total runtime is heavily dominated by the solution of the system of linear equations using the \gls{cg} algorithm if the data set is sufficiently large enough.
For data sets with less than $2^{12}$ data points, the IO components \enquote{read} and \enquote{write} contribute more to the total runtime than solving the system of linear equations.
However, for a data set size of $2^{15}$, the \gls{cg} algorithm is responsible for $\SI{92}{\percent}$ of the total runtime, the other measured components combined are only accountable for $\SI{5}{\percent}$ and, therefore, negligible.
The remaining $\SI{3}{\percent}$ of the runtime are due to parts that are not displayed here, such as the overhead when accessing the \gls{gpu} for the first time or the cleanup at the end of the program execution.
\Autoref{fig:components:datapoints} shows that \enquote{read}, \enquote{transform}, and \enquote{write} do scale better than \enquote{cg} for an increasing number of data points. 
They can therefore also be neglected for the runtime of even larger training sets.
The \enquote{cg} component has the worst complexity: doubling the number of data points increases the runtime by a factor of $\num{3.3}$.

In \autoref{fig:components:features} we proceed as before and fix the number of data points to $2^{15}$ while varying the number of features from $2^{6}$ to $2^{14}$.
The result stays the same: solving the system of linear equations is again responsible for over $\SI{92}{\percent}$ of the total runtime.
Doubling the number of features increases the runtime by roughly a factor of $\num{2.11}$.
Here, a factor of two can be justified by the fact that the effort for implicitly calculating each matrix entry doubles, since the vectors used in the scalar products have twice the size.
Furthermore, the problem becomes more complex in higher dimension, so that it can be observed that more \gls{cg} iterations are needed to reach a similar accuracy.
Again, this factor is problem dependent.
In any case, a factor larger than two is to be expected, provided that the additional features actually lift the problem into a higher dimension.
First tests have indicated that the factor is close to two if the vectors are only extended with zeros.

\subsection{Runtime and Accuracy Depending on Epsilon}
\label{sec:runtime_acc_eps}

\begin{figure}
	\centering
	\subfloat[Runtime and number of \gls{cg} iterations relative to the relative residual.\label{fig:eps:datapoints}]{\includegraphics[width=.49\linewidth]{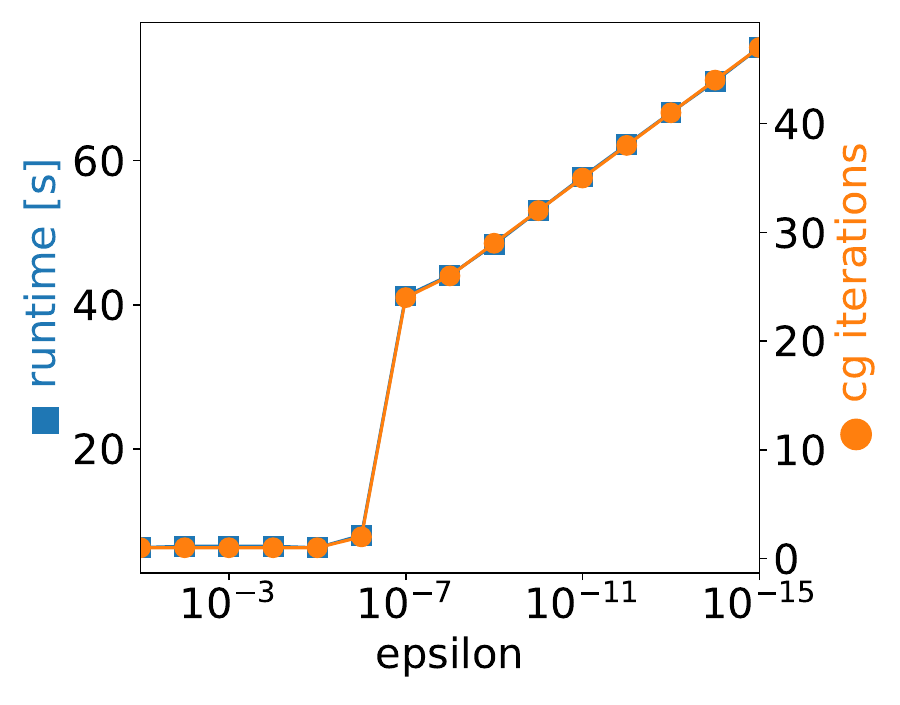}}
	\hfill
	\subfloat[Accuracy and number of \gls{cg} iterations relative to the relative residual.\label{fig:eps:features}]{\includegraphics[width=.49\linewidth]{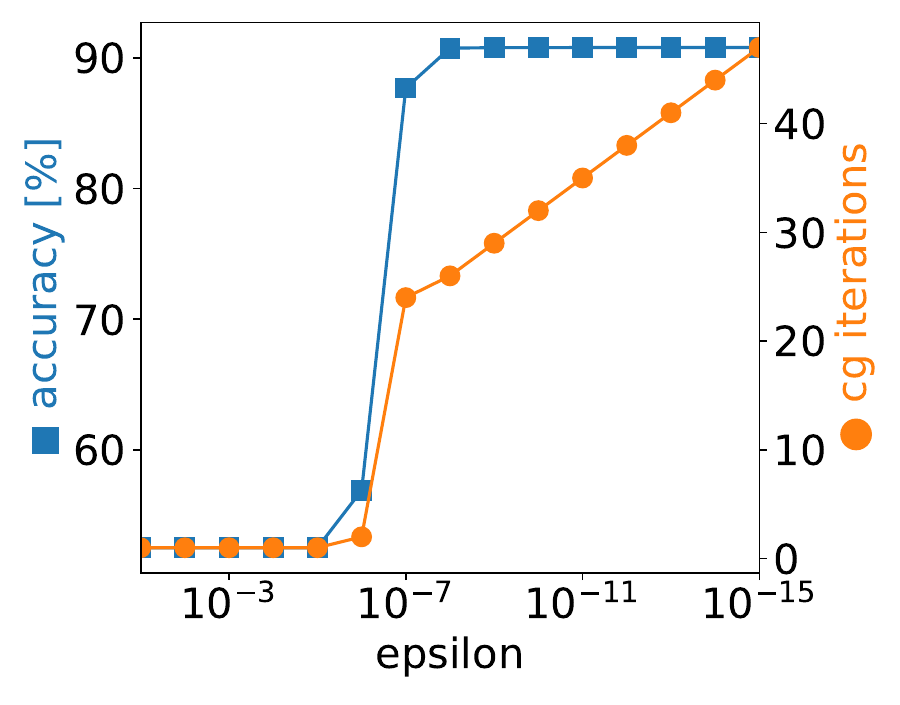}}
	\caption{Runtime, accuracy, and number of \gls{cg} iterations in relation to the relative residual of the \gls{cg} method's epsilon. 
	Measured on the \gls{gpu} machine with $2^{15}$ data points and $2^{12}$ features each.}
	\label{fig:eps}
	\vspace*{-1em}
\end{figure}

\Autoref{fig:eps} displays the runtime and achieved accuracy based on the chosen epsilon using a single \gls{gpu}. 
The epsilon is used as a termination criterion in the \gls{cg} algorithm.
The data set contains $2^{15}$ data points with $2^{12}$ features each.

\Autoref{fig:eps:datapoints} shows the runtime and number of \gls{cg} iterations with respect to epsilon.
For epsilon values up to 1e-06, the number of \gls{cg} iterations does not increase significantly.
Refining the epsilon $\times0.1$ further, increases the number of \gls{cg} iterations by a factor of twelve from $2$ to $24$.
After that, refining the epsilon even further, steadily increases the number of \gls{cg} iterations on average by $2$.
As already discussed in \autoref{sec:components}, the total runtime is heavily dominated by solving the system of linear equations.
This in turn is dominated by the number of \gls{cg} iterations.
Therefore, both lines lie on top of each other. 
This is additionally underlined by the fact that the runtime per \gls{cg} iteration stays the same for data sets of the same size.

A different behavior can be observed in \autoref{fig:eps:features}.
Here, the accuracy and number of \gls{cg} iterations correlate until an epsilon of 1e-05 is reached.
Afterward, the accuracy jumps to $\SI{56.9}{\percent}$ followed by $\SI{87.6}{\percent}$ and then reaching its final value of $\SI{90.8}{\percent}$ for an epsilon of 1e-08. 
However, the number of \gls{cg} iterations steadily grows from $24$ iterations for an epsilon of 1e-07 to $47$ iterations for an epsilon of 1e-15.

In general, it is nice to note that the runtime does not explode when decreasing epsilon by eight orders of magnitude from 1e-07 to 1e-15, it merely grows by a factor of about $\num{1.83}$.
Thus, if a high accuracy is desired, it is fine to select a relatively small epsilon; the exact choice is not critical.

\subsection{Scaling on a Many-Core CPU and Multiple GPUs}
\label{sec:scaling}

\begin{figure}
	\centering
	\subfloat[Scaling behavior of the different \gls{plssvm} components with respect to the number of cores for $2^{12}$ data points and $2^{11}$ features.\label{fig:scaling:cpu}]{\includegraphics[width=.45\linewidth]{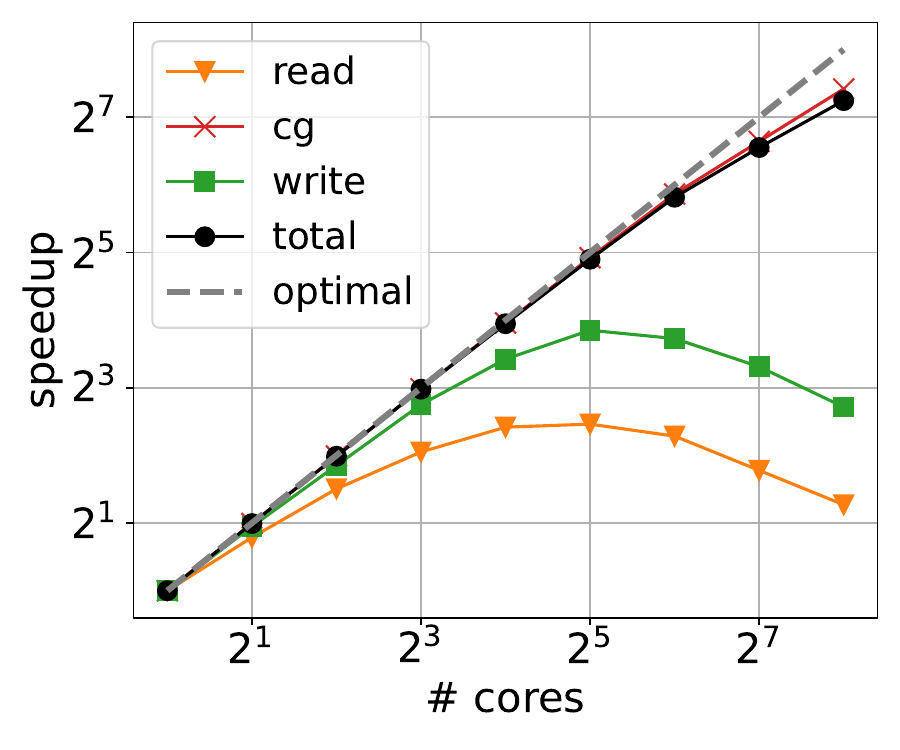}}
	\hfill
	\subfloat[Scaling behavior of the different \gls{plssvm} components with respect to the number of \glspl{gpu} for $2^{16}$ data points and $2^{14}$ features.\label{fig:scaling:gpu}]{\includegraphics[width=.45\linewidth]{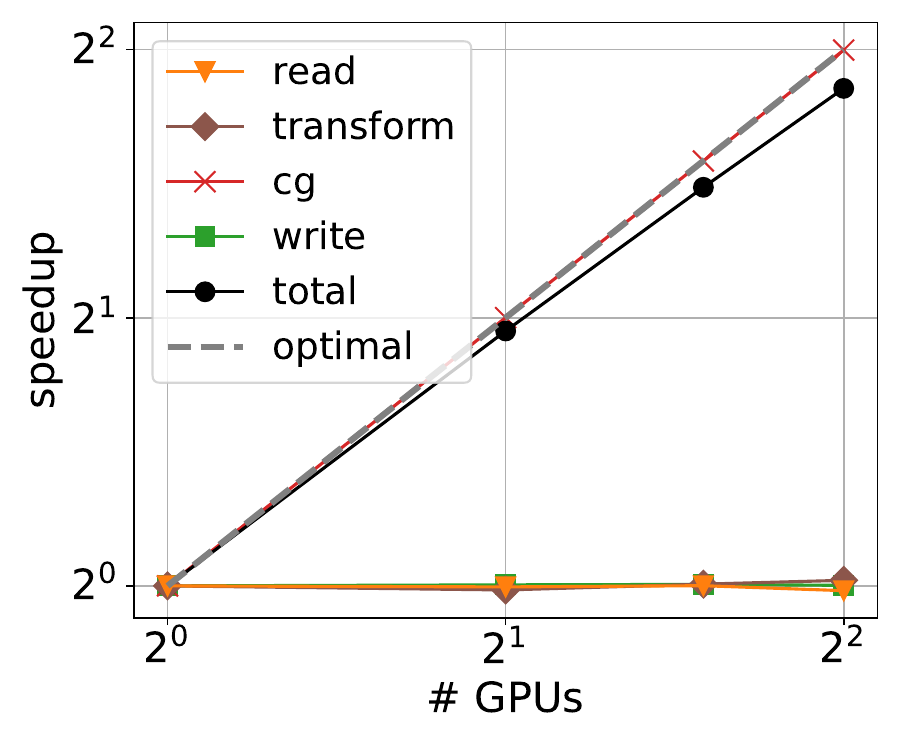}}
	\caption{Scaling behavior with an increasing number of available hardware for the \gls{plssvm} components on a many-core \gls{cpu} and a multi-\gls{gpu} system.}
	\label{fig:scaling}
	\vspace*{-1em}
\end{figure}

\Autoref{fig:scaling} shows the scaling behavior, displayed in a double-logarithmic graph, for the different \gls{plssvm} components with respect to the number of cores on a \gls{cpu} and the number of \glspl{gpu} in the system.
On the \gls{cpu}, only the linear kernel is plotted, since the overall scaling behaviors of the polynomial and radial kernels are the same.
For the multi-\gls{gpu} runs, we were only able to investigate the linear kernel, since the polynomial and radial kernels do not yet support the execution on multiple \glspl{gpu}.

\Autoref{fig:scaling:cpu} shows the speedup for the \enquote{read}, \enquote{cg}, and \enquote{write} components on a \gls{cpu} with $2 \cdot 64$ physical cores and $2 \cdot 128$ hyper-threads using the OpenMP backend.
The \enquote{transform} component is omitted since the $2D$ to $1D$ transformation is only applied for the \gls{gpu} backends.
The classification data set contains $2^{12}$ data points with $2^{11}$ features.

We observe that all components of our \gls{plssvm} library scale equally well with an increasing number of available \gls{cpu} cores on up to $16$ cores.
The runtime decreases from $\SI{25.3}{\minute}$ on a single core to $\SI{3.1}{\minute}$ on $16$ cores.
Using more than $64$ cores increases the runtime of the \enquote{read} and \enquote{write} components: At this point, OpenMP switches from a single socket to two sockets on our system.
However, the \enquote{cg} component scales well even for up to $256$ \gls{cpu} cores.
In this case, the initial runtime of $\SI{25.3}{\minute}$ has been reduced to $\SI{20}{\second}$, which corresponds to a parallel speedup of roughly $\num{74.7}$.

\Autoref{fig:scaling:gpu} shows the scaling behavior of the same three components together with the \enquote{transform} component on up to four A100 \glspl{gpu} using the \gls{cuda} backend.
Here, the data set contains $2^{16}$ data points with $2^{14}$ features.
Increasing the number of \glspl{gpu} does not result in better runtimes for the \enquote{read}, \enquote{transform}, and \enquote{write} components. 
This is rather obvious, since none of them uses the \glspl{gpu}.
The \enquote{cg} component, which dominates the total runtime, is executed on the \glspl{gpu}.
Using four \glspl{gpu} reduces the total runtime by a factor of $\num{3.71}$ from $\SI{13.49}{\minute}$ to $\SI{3.72}{\minute}$. 
This nicely demonstrates good scaling on multiple \glspl{gpu}.

The total amount of memory used for the given data set on a single \gls{gpu} is $\SI{8.15}{\gibi\byte}$.
Using four \glspl{gpu} reduces the amount of memory used to $\SI{2.14}{\gibi\byte}$ per \gls{gpu}.
While this results only in a reduction of a factor of $\num{3.6}$, instead of the optimal factor of $\num{4}$, it nevertheless shows that using multiple \glspl{gpu} not only allows us to train \gls{svm} classifiers more efficient, but also to process data sets that would not fit on a single \gls{gpu}.
In contrast, using the same data set with ThunderSVM results in a memory consumption of $\SI{13.08}{\gibi\byte}$ on a single \gls{gpu}.


\subsection{Comparison of the Implementations}
\label{sec:comparison}

To summarize, all four implementations show competitive performance for solving binary classification problems on the \gls{cpu}. 
\gls{plssvm} and ThunderSVM can utilize modern \glspl{gpu}.
ThunderSVM is restricted to \glspl{gpu} from NVIDIA, as its backend is exclusively written in CUDA. \gls{plssvm} is the only implementation supporting more than one \gls{gpu} as well as \glspl{gpu} from different vendors.

Since LIBSVM and ThunderSVM are well-established \gls{svm} libraries, they provide more functionality than \gls{plssvm} so far. 
For example, all libraries have specific implementations for linear, polynomial, and radial basis functions kernels. 
In addition, LIBSVM and ThunderSVM have a sigmoid kernel and LIBSVM even supports precomputed kernels. 
Multi-class classification as well as regression are not yet supported by our \gls{plssvm} library, in contrast to ThunderSVM and LIBSVM. 
However, it is not difficult to include these functionalities on the basis of our library if needed.

Note that the state-of-the-art \gls{smo} algorithm and the least squares approach solve slightly different optimization problems.
Therefore, for a given data set, the termination criterion (epsilon) can not be reused between the two methods.
This makes comparing the overall runtimes between the two approaches more difficult.
In general, straightforward problems are better suited for \gls{smo}, as only few support vectors can be sufficient to compute the separating hyperplane. 
Difficult classification problems with plenty of data points and complex clusters that require many support vectors are better suited for the \gls{lssvm} approach.
The termination criterion for the iterative solver of our \gls{plssvm} has to be selected suitably. 
However, all of our classification tasks so far have shown that the runtime is not very sensitive with respect to the termination criterion epsilon for high accuracies.

Finally, both ThunderSVM and LIBSVM support sparse data representations, which are used in their internal calculations.
\gls{plssvm} so far only supports sparse data representations when reading and writing. 
When parsing sparse data, we allocate memory for all features including those that are zero, resulting in a dense data representation internally.


\section{Conclusions and Future Work}
\label{sec:conclusion}

In this work, we introduced the new \gls{svm} library \gls{plssvm}.
It is based on the rarely used least squares approach.
In contrast to the \gls{smo} method that is used in most state-of-the-art implementations, it includes all data points as support vectors to compute the class-separating hyperplane.
While \gls{smo} has limited parallelization potential, the \gls{lssvm} approach is well-suited for massively parallel hardware and thus large data sets.
\gls{plssvm} is the very first \gls{svm} implementation supporting multiple backends---OpenMP, \gls{cuda}, \gls{opencl}, and SYCL---to be able to target different hardware platforms from various vendors, being it \glspl{cpu} or \glspl{gpu}.

We demonstrated with classification tasks for artificial and real-world dense data sets of different sizes that we are capable of competing with well-established \gls{smo} implementations such as LIBSVM or ThunderSVM on the \gls{cpu} although our basic implementation is currently still very much open to optimizations.
Since there are already many widely distributed implicit matrix-vector multiplication implementations available that we have not considered yet, this shows that there is much potential for further work on a highly scalable multi-node \gls{lssvm} implementation.
On the \gls{gpu} our library shows its main strength and severely outperforms ThunderSVM by a factor of $14$. 
We verified that our implementation scales well on many-core \glspl{cpu} and multiple \glspl{gpu} achieving a parallel speedup of $\num{3.71}$ on four NVIDIA A100 \glspl{gpu}. 

The \gls{gpu} implementations have a small overhead accessing the \gls{gpu}(s).
Therefore, the \gls{cpu} implementations are better suited for learning classifiers for very small data sets.
Note that \gls{plssvm} is currently implemented using a dense \gls{cg} implementation.
In the case of very sparse data sets with many features, it is therefore better to use ThunderSVM. 
\gls{plssvm} clearly outperforms both LIBSVM and ThunderSVM for large non-trivial classification tasks by orders of magnitude, making use of the high parallelization potential of \glspl{lssvm}.

Canonical next steps include the optimization of the \gls{cpu} implementation, to consider sparse data structures for the \gls{cg} solver, and to target multi-node multi-\gls{gpu} systems to be able to use even larger data sets as currently possible and to report the scaling behavior of our library on more \glspl{gpu} and \gls{cpu} cores.
In future work, we will investigate the advantages and disadvantages of the different backends, and provide extensive scalability studies on non-NVIDIA \glspl{gpu}. 
As a long-term future task, we want to extend all \gls{plssvm} kernels to support multi-node multi-\gls{gpu} execution including load balancing on heterogeneous hardware. 
Finally, we intend to extend \gls{plssvm} to provide all the standard functionality of LIBSVM to our users. 
This includes multi-class classifications and regression tasks.

\section*{Acknowledgement}
\addcontentsline{toc}{section}{Acknowledgments}
We thank the Deutsche Forschungsgemeinschaft (DFG, German Research Foundation) for supporting this work by funding -- EXC2075 -- 390740016 under Germany's Excellence Strategy. We acknowledge the support by the Stuttgart Center for Simulation Science (SimTech).

\bibliography{IEEEabrv, Literature/bibliography}
\end{document}